\documentclass[a4paper,UKenglish,cleveref,autoref,pdfa]{lipics-v2021}
\hideLIPIcs

\nolinenumbers

\hideLIPIcs  

\title{Towards a Model of Puzznic}

\author{Joan Espasa}{School of Computer Science, University of St Andrews, UK}{jea20@st-andrews.ac.uk}{https://orcid.org/0000-0002-9021-3047}{}
\author{Ian P. Gent}{School of Computer Science, University of St Andrews, UK}{ian.gent@st-andrews.ac.uk}{https://orcid.org/0000-0002-5604-7006}{}
\author{Ian Miguel}{School of Computer Science, University of St Andrews, UK}{ijm@st-andrews.ac.uk}{https://orcid.org/0000-0002-6930-2686}{EPSRC grant EP/V027182/1}
\author{Peter Nightingale}{Department of Computer Science, University of York, UK}{peter.nightingale@york.ac.uk}{https://orcid.org/0000-0002-5052-8634}{EPSRC grant EP/W001977/1}
\author{Andr\'as Z. Salamon}{School of Computer Science, University of St Andrews, UK}{andras.salamon@st-andrews.ac.uk}{https://orcid.org/0000-0002-1415-9712}{}
\author{Mateu Villaret}{Department of Computer Science, Applied Mathematics and Statistics, University of Girona, Spain}{mateu.villaret@udg.edu}{https://orcid.org/0000-0002-8066-3458}{Grant PID2021-122274OB-I00 funded by MCIN/AEI/10.13039/501100011033 and by ERDF A way of making
  Europe}

\authorrunning{Espasa et al.} 

\Copyright{Joan Espasa and Ian P. Gent and Ian Miguel and Peter Nightingale and András Z. Salamon and Mateu Villaret} 

\ccsdesc[500]{Theory of computation~Constraint and logic programming}
\ccsdesc[500]{Computing methodologies~Planning and scheduling}

\keywords{AI Planning, Modelling, Constraint Programming} 

\category{} 

\relatedversion{} 

\acknowledgements{We thank Emily Miguel for an empirical investigation of Puzznic's mechanics.}

\EventEditors{John Q. Open and Joan R. Access}
\EventNoEds{2}
\EventLongTitle{42nd Conference on Very Important Topics (CVIT 2016)}
\EventShortTitle{CVIT 2016}
\EventAcronym{CVIT}
\EventYear{2016}
\EventDate{December 24--27, 2016}
\EventLocation{Little Whinging, United Kingdom}
\EventLogo{}
\SeriesVolume{42}
\ArticleNo{23}

\definecolor{mygreen}{rgb}{0,0.6,0}

\usepackage{listings}
\lstdefinelanguage{PDDL}{
  backgroundcolor=\color{black!5},
  keywords={and, or, exists, forall, assign, not, when},
  keywordstyle=\color{blue}\bfseries,
  ndkeywords={action, parameters, precondition, effect, goal, derived},
  ndkeywordstyle=\color{mygreen}\bfseries,
  identifierstyle=\color{black},
  sensitive=false,
  comment=[l]{;},
  commentstyle=\color{purple}\ttfamily,
  stringstyle=\color{red}\ttfamily,
  morestring=[b]',
  morestring=[b]"
}

\lstdefinelanguage{eprime}{
  backgroundcolor=\color{black!5},
  keywords={forAll, or, and, letting, union, max, sum, exists, int, of, find, given, flatten, matrix, indexed, by, be, domain, ->, /\\},
  keywordstyle=\color{blue}\bfseries,
  ndkeywords={atleast, atmost, gcc},
  ndkeywordstyle=\color{mygreen}\bfseries,
  identifierstyle=\color{black},
  sensitive=false,
  comment=[l]{\$},
  commentstyle=\color{purple}\ttfamily,
  stringstyle=\color{red}\ttfamily,
  morestring=[b]',
  morestring=[b]",
}

\lstdefinelanguage{essence}{
 backgroundcolor=\color{black!5},
 frame = single,
 breaklines=true,
 keywords = { language, Essence, given, letting, find, such, that, domain, function, total, surjective, be , forAll, exists, injective, in, preImage, range ,  mset, set, partition, new, type, intersect, from , minimising, maximising, indexed, by, defined, maxSize, maxNumParts , subset , size, toInt, sum , sequence },
  keywordstyle=\color{blue}\bfseries,
  ndkeywords={atleast, atmost, gcc, int, matrix, bool},
  ndkeywordstyle=\color{mygreen}\bfseries,
  identifierstyle=\color{black},
  sensitive=false,
  comment=[l]{\$},
  commentstyle=\color{purple}\ttfamily,
  stringstyle=\color{red}\ttfamily,
  morestring=[b]',
  morestring=[b]",
}

\usepackage[normalem]{ulem}
\usepackage{xspace}

\newcommand{\eprime}{{\sc Essence Prime}\xspace}
\newcommand{\savilerow}{{Savile Row}\xspace}
\newcommand{\conjure}{\textsc{Conjure}\xspace}
\newcommand{\essence}{\textsc{Essence}\xspace}

\newcommand{\joan}[1]{\color{orange}{#1}\color{black}}

\newcommand{\zap}[1]{} 
\newcommand{\code}[1]{{\texttt{#1}}}

\begin{document}

\maketitle

\begin{abstract}
We report on progress in modelling and solving Puzznic, a video game requiring the player to plan sequences of moves to clear a grid by matching blocks.
We focus here on levels with no moving blocks.
We compare a planning approach and three constraint programming approaches on a small set of benchmark instances. The planning approach is at present superior to the constraint programming approaches, but we outline proposals for improving the constraint models.
\end{abstract}

\section{Introduction}

\begin{figure}[t!] 
\centering
\includegraphics[scale=0.3]{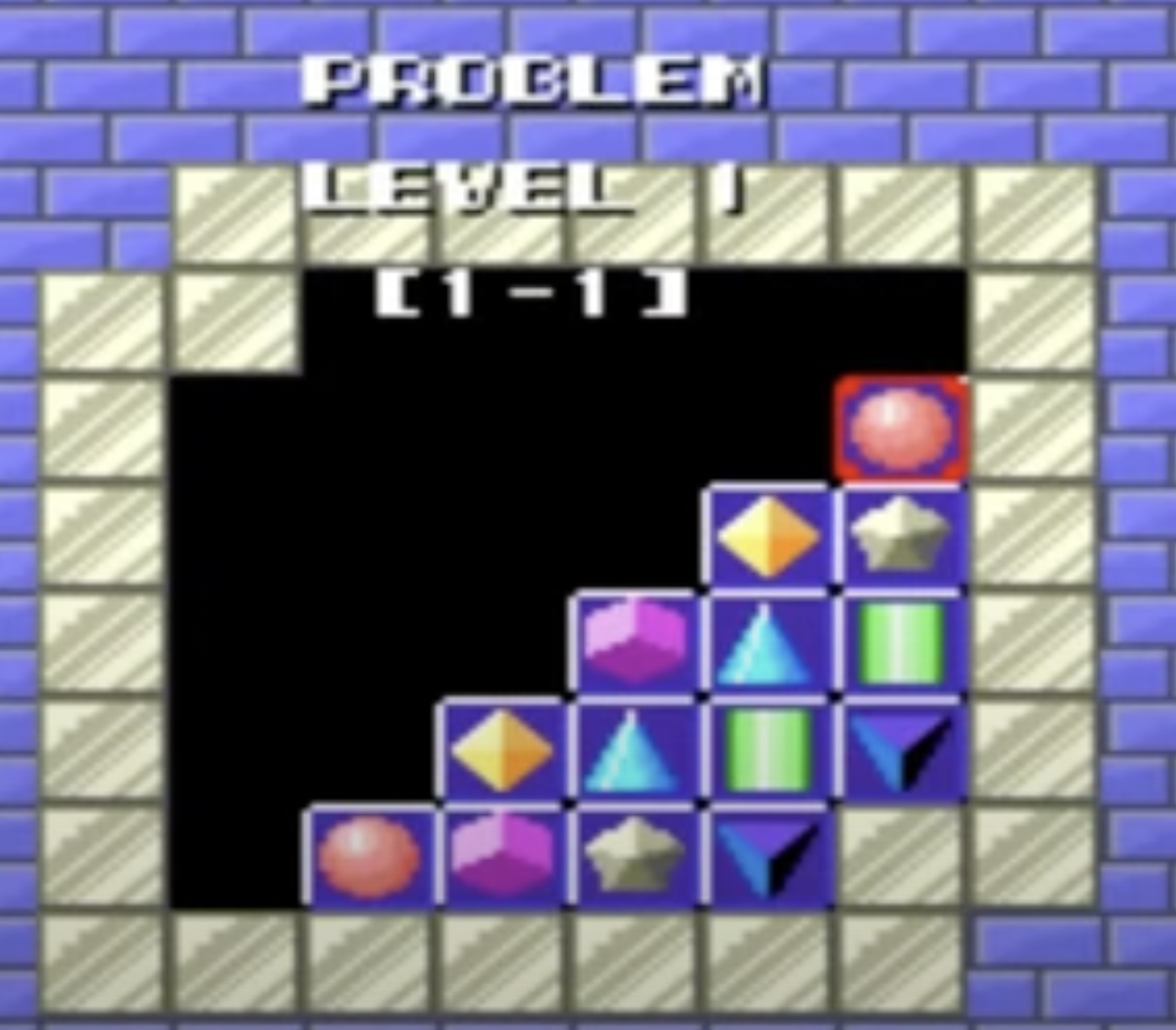}
\caption{Detail from Puzznic (Taito, 1989). The red cursor that can be seen on the red circle block at the upper right of the image is controlled by the player and used to select and move individual blocks. When similarly-patterned blocks are adjacent horizontally or vertically they will match and are removed from play. The goal is to match and remove all patterned blocks in the level.}
\label{fig:screenshot}
\end{figure}

We focus on modelling and solving {\em Puzznic}, a puzzle-based video game published by Taito in 1989 and ported to many platforms. The player manipulates blocks in a given grid until they {\em match} when two or more blocks of the same pattern are adjacent, and are removed from play. The goal is to match, and so remove, all patterned blocks in the grid. An illustrative level from the game is presented in Figure \ref{fig:screenshot}.

Puzznic is naturally characterised as a Planning problem, one of the fundamental disciplines of Artificial Intelligence \cite{ghallab2004automated}. Given a model of the environment (here the grid, blocks, and their behaviour), a planning problem requires finding a sequence of actions (block moves made by the player) to progress from an initial state of the environment to a goal state (all blocks matched) while respecting a set of constraints.

Constraint Programming has been used to solve planning problems \cite{bartak2010constraint,related1} and has recently proven successful in solving Plotting \cite{Espasa2022:plotting}, a similar puzzle game which, in common with Puzznic, has complex changes of state. The full version of Puzznic has dynamic elements in the form of moving blocks. Herein, we focus on the core of the game without these dynamic elements and demonstrate how it can be modelled successfully both in the constraints paradigm and with PDDL, the standard modelling language of the AI Planning community.

We discuss both Planning and Constraint approaches.  One general point that we have \emph{not} focussed on is having a uniform measure of cost of a solution.  Each different approach has a natural measure of cost, and we optimise each with respect to its own measure.  This does mean that plan lengths from one measure are not necessarily optimal with respect to another measure.  On the other hand, this allows each method to be tuned to its best advantage so is unlikely to disadvantage any one method artificially. 

\section{Puzznic}

Puzznic is a puzzle solitaire game, in which the player has full information of the state of the game, and the effects of each action performed are deterministic. This is representative of a variety of puzzle video games, such as the aforementioned Plotting, as well as board games like peg solitaire~\cite{Jefferson2006:modelling} and some variants of patience like Black Hole~\cite{Gent2007:search}.

Each instance of the game consists of a grid of cells similar to that presented in Figure~\ref{fig:screenshot}. Each cell may be empty or filled with a wall or contain a patterned  block. The player controls a red cursor, visible at the top-right of the figure, with which they can select a single patterned block. Once selected, a block can be moved horizontally left or right, but only if the cell in the direction chosen is empty. Patterned blocks are affected by gravity, and will fall until coming to rest above another non-empty block or a wall. If the player moves a block over an empty cell, they lose control of the block as it falls.

When two or more  blocks with the same pattern are adjacent horizontally or vertically, they {\em match} and are removed from play. The only exception to this is when a patterned block is falling: it cannot match another block until it comes to rest. Via gravity, one match may result in further matches, and so on. Such cascading matches are rewarded by the scoring system. The goal is to match and remove all patterned blocks from the grid.

Some levels of the game also have moving wall blocks, which can carry patterned blocks. In the current work we do not consider this aspect of Puzznic. The version of the game studied here, without moving wall blocks, is also known as \emph{Cubic} and the solvability of this simpler variant has been claimed to be NP-hard by a reduction from deciding satisfiability of Boolean circuits~\cite{cubicnp}. The complexity of the related Hanano puzzle has also been studied~\cite{Liu2019:hanano}, and recently this problem was shown to be \textsf{PSPACE}-complete~\cite{chavrimootoo2022:defying}.


\section{A PDDL Formulation}
\label{section:pddl}

Considering Puzznic as a classical planning problem requires finding a sequence of actions (a \emph{plan}) where their application will successively transform a given initial state until a goal state is reached. 
A set of finite-domain variables determines the state at each moment. An action is applicable at a certain state if the state satisfies the preconditions of the action. The state is modified according to the effects of the action.

The Planning Domain Definition Language (PDDL)~\cite{pddl} is the \emph{de facto} planning definition standard. PDDL separates a planning problem into two files: the \emph{domain}, defining general characteristics of the problem such as the representation of the state and how the actions operate, and the \emph{problem}, which defines the objects, the initial state and the goal of a particular instance.
In this section we describe our PDDL Puzznic formulation.

\subsection*{The state}

We only consider objects of the following types: \emph{location} to represent a grid cell location; the \code{up}, \code{down}, \code{left} and \code{right} \emph{direction}s used both to relate locations and to specify movements; and finally \emph{pattern} to represent the patterns of blocks used in the scenario. 

Notice that we do not define a block object in the domain. Blocks are instead represented by patterns assigned to locations.
If a location has no pattern this means that no block is there, while if a location has a certain pattern then there is a block of that pattern there.

In order to model the state, the following predicates are defined:
\begin{lstlisting}[escapechar=|,language=PDDL,basicstyle=\scriptsize\ttfamily,columns=fullflexible,keepspaces=true,tabsize=2,stepnumber=1]
  (patterned ?l - location ?p - pattern)
  (next ?from ?to - location ?dir - direction)
  (free ?l - location)
  (falling_flag)
  (matching_flag)
\end{lstlisting}
The {\tt patterned} predicate allows us to state what is the pattern of a certain location. With the {\tt next} predicate we state what location we find following a certain direction from a location. We treat the \code{next} predicate as a declarative specification of the game grid adjacency graph.
Further, because we can never move wall blocks, we exclude walls from the grid.
This choice to only model the non-wall locations in the game grid also leads to a smaller state space.
With the {\tt free} predicate we state whether a certain location is free or not.
We also use two ``flag'' predicates to represent 
gravity and block matching semantics of the game.

The \code{free}, \code{falling\_flag} and \code{matching\_flag} predicates are \emph{derived predicates}, which are automatically updated after the application of each action:

\begin{lstlisting}[escapechar=|,language=PDDL,basicstyle=\scriptsize\ttfamily,columns=fullflexible,keepspaces=true,tabsize=2,stepnumber=1]
; a block is free if it is not patterned
 (:derived (free ?l)
    ( forall (?p - pattern)  (not (patterned ?l ?p)) ))

; is there something that needs to fall?
 (:derived (falling_flag)
    ( exists (?l1 ?l2 - location) (and (next ?l1 ?l2 down) (not (free ?l1)) (free ?l2)) ))

; is there something that needs to match?
 (:derived (matching_flag)
    ( exists (?l1 ?l2 - location ?p - pattern ?d - direction)
        (and (next ?l1 ?l2 ?d) (patterned ?l1 ?p) (patterned ?l2 ?p)) ))
\end{lstlisting}

\subsection*{Actions}

Actions are defined by their parameters as well as by their preconditions and effects which usually constrain the parameters. Preconditions define the requirements a state must satisfy in order for the action to be applicable. Effects define how actions change the state once an action has been applied. 
Three actions are defined: \code{move\_block}, \code{fall\_block} and \code{match\_blocks}. The solving process of the planner needs to follow the semantics of the game, which can be concisely summarized with this solving algorithm:
\begin{verbatim}
while there are blocks remaining do
    if     falling_flag:  fall_block
    elseif matching_flag: match_blocks
    else   move_block
\end{verbatim}

\zap{
\begin{algorithmic}
\While{\emph{there are blocks remaining}}
    \If{\code{falling\_flag}}
        \State \text{\code{fall\_block}}
    \ElsIf{\code{matching\_flag}}
             \State \text{\code{match\_blocks}}
    \Else
        \State \text{\code{move\_block}}
    \EndIf
\EndWhile
\end{algorithmic}
}

PDDL allows us to concisely express the three needed actions, which are presented as follows. We use the predicate flags in the preconditions to control when each action can be applied, according to the above algorithm.

First, the falling action:

\begin{lstlisting}[escapechar=|,language=PDDL,basicstyle=\scriptsize\ttfamily,columns=fullflexible,keepspaces=true,tabsize=2,stepnumber=1]
(:action fall_block
  :parameters
  (?l1 ?l2 - location ?p - pattern)
  :precondition 
  (and
    (falling_flag)      ; something needs to fall
    (next ?l1 ?l2 down) ; l1 is on top of l2
    (patterned ?l1 ?p)  ; l1 has some pattern and needs to fall
    (free ?l2)          ; l2 is free as we're falling on it
  )
  :effect 
  (and
    ; the patterns get properly assigned: l1 loses the pattern and l2 gains the pattern l1 had
    (not (patterned ?l1 ?p)) (patterned ?l2 ?p)
  ))
\end{lstlisting}

Next, the matching action:

\begin{lstlisting}[escapechar=|,language=PDDL,basicstyle=\scriptsize\ttfamily,columns=fullflexible,keepspaces=true,tabsize=2,stepnumber=1]
 (:action match_blocks
  :parameters ()
  :precondition 
  (and
    ; first things fall, then they match
    (not (falling_flag))
    (matching_flag)
  )
  :effect 
  (and
    (forall (?l1 - location ?p - pattern)
        (when  ; if a patterned locations has some neighbor with the same pattern
          ( exists (?l2 - location ?d - direction)
            (and (next ?l1 ?l2 ?d) (patterned ?l1 ?p) (patterned ?l2 ?p)) )
          (not (patterned ?l1 ?p)) ; remove its pattern
        )))))
\end{lstlisting}

Finally, the moving action:

\begin{lstlisting}[escapechar=|,language=PDDL,basicstyle=\scriptsize\ttfamily,columns=fullflexible,keepspaces=true,tabsize=2,stepnumber=1]
(:action move_block
  :parameters
  (?l ?tl - location ?d - direction ?p - pattern)
  ;; ?tl - target location where we move the block to
  ;; ?l - original place of the block we're moving
  ;; ?d - the direction we're moving the block
  ;; ?p - the pattern of the block we move
  :precondition
  (and  ; if nothing falls and nothing matches we can move
    (not (falling_flag)) (not (matching_flag))
    (or (= ?d right) (= ?d left))   ; only left/right movements allowed
    (patterned ?l ?p)               ; ?l has the ?p pattern
    (next ?l ?tl ?d) (free ?tl)     ; the target location where we want to move must be free
    )
   :effect
    (and
        (not (patterned ?l ?p)) (patterned ?tl ?p)  ; update patterns of the locations
        (increase (total-cost) 1)                   ; this is a non-free action
    ))                 
\end{lstlisting}

\subsection*{The Goal}

As the goal is to remove all patterned blocks from the grid, we state that we want to reach a state where no location has a pattern, additionally asking for the minimum number of moves.

\begin{lstlisting}[escapechar=|,language=PDDL,basicstyle=\scriptsize\ttfamily,columns=fullflexible,keepspaces=true,tabsize=2,stepnumber=1]
    (:goal  ( forall (?l - location) (not (exists (?p - pattern) (patterned ?l ?p))) ))
    (:metric minimize (total-cost))
\end{lstlisting}



\section{Constraint Models in \texorpdfstring{\eprime}{Essence Prime}}
\label{section:constraints}

To implement Puzznic via constraints, we based our models on steps. Each step is either a move of a cell by the player, or matched cells being removed.   In both cases gravity is automatically applied instantaneously, which represents the major change from the PDDL model described in \cref{section:pddl}.
We made two variants of the model.  The first requires an exact fixed number of steps, with no missing steps.  The second is based on player moves, so does not count the total matching steps.  It also differs by not requiring the bound on the number of moves to be met exactly.
We describe the two models in the following subsections, but only describe ways in which the second model differs from the first since it is similar.

We wish to draw attention to one feature of our constraint models.
We use the declarative nature of constraints to avoid having to explicitly calculate the effects of gravity when blocks disappear due to matching.  For patterned blocks in a column affected by matching, we insist that they stay in the same order, and that none end in a position that is above an empty cell.  From this the constraint model can deduce what position each block will arrive in, without the constraint modeller having to add constraints to calculate the position explicitly.

\subsection{Fixed Steps \texorpdfstring{\eprime}{Essence Prime} Model}
\label{section:fixedsteps}

Our constraint model is formulated in \eprime \cite{savile-row-manual-caps}. We begin by describing the parameters to the model and defining some useful constants:

\begin{lstlisting}[escapechar=|, language=eprime]
letting WALL be 0
letting EMPTY be 1
given initGrid : matrix indexed by[int(1..gridHeight), int(1..gridWidth)] of int(0..)
letting GRIDCOLS be domain int(1..gridWidth)
letting INTERIORCOLS be domain int(2..gridWidth-1)
letting GRIDROWS be domain int(1..gridHeight)
letting INTERIORROWS be domain int(2..gridHeight-1)
letting PATTERNS be domain int(2..max(flatten(initGrid)))

given noSteps : int(1..)
letting STEPSFROM1 be domain int(1..noSteps)
letting STEPSFROM0 be domain int(0..noSteps)
letting STEPSEXCEPTLAST be domain int(0..noSteps-1)
\end{lstlisting}

The first parameter to the model is \code{initGrid}, which captures in a two-dimensional matrix the initial state. Possible values for \code{initGrid} range from 0 (which we use to represent a wall block) upwards, with 1 representing an empty cell, and the remaining values denoting each of the patterned block types in the initial grid. It is assumed that \code{initGrid} has a perimeter of wall blocks. Non-rectangular levels are represented by creating a large enough bounding rectangle and using wall blocks for the unused cells.

The second parameter specifies the number of steps for which a plan is sought. In common with many constraint models of planning problems, for example in solving Plotting \cite{Espasa2022:plotting}, we solve a sequence of decision problems of increasing \code{noSteps}, such that the first such instance for which a solution can be found provides the optimal length plan.

\subsubsection{Viewpoint}

Our constraint model follows a common pattern in constraint models of AI planning problems in employing a time-indexed set of variables, interleaving an account of the state of the puzzle with the action taken to transform the previous state into that following. We consider step 0 to be the initial state. The action taken in step 1 then modifies the initial state to produce the state in step 1, and so on. A key part of the viewpoint of the model is therefore these state and action variables:

\begin{lstlisting}[escapechar=|, language=eprime]
find moveRow : matrix indexed by[STEPSFROM1] of INTERIORROWS
find moveCol : matrix indexed by[STEPSFROM1] of INTERIORCOLS
find moveDir : matrix indexed by[STEPSFROM1] of int(-1,1)
find grid : matrix indexed by[STEPSFROM0, GRIDROWS, GRIDCOLS] of int(WALL) union int(EMPTY) union PATTERNS
\end{lstlisting}

The player's action is captured by \code{moveRow} and \code{moveCol}, which indicate the block selected to move, and \code{moveDir}, which indicates whether the block is moved left (via the value -1) or right (via the value 1). For simplicity, at present we consider only movement of a single cell per time step. In future we will investigate supporting multiple cell moves in a single step.

We also introduce auxiliary variables to record the destination of the moved block:
\begin{lstlisting}[escapechar=|, language=eprime]
find destRow : matrix indexed by[STEPSFROM1] of INTERIORROWS
find destCol : matrix indexed by[STEPSFROM1] of INTERIORCOLS
\end{lstlisting}
These simplify some of the problem constraints, and in particular the handling of gravity.

Puzznic has relatively complex changes of state, due both to gravity and matching (multiple simultaneous matches are possible). For this reason our constraint model treats matching as a separate step in the plan in which the player makes no action. In particular this facilitates dealing with the effects of cascaded matches. We introduce a Boolean variable to indicate which mode, matching or player action, is in effect:
\begin{lstlisting}[escapechar=|, language=eprime]
find matching : matrix indexed by[STEPSFROM1] of bool
\end{lstlisting}

Finally, we introduce auxiliary variables to help detect matching and process its effects:
\begin{lstlisting}[escapechar=|, language=eprime]
find matchingGrid : matrix indexed by[STEPSEXCEPTLAST, INTERIORROWS, INTERIORCOLS] of bool

find mapsToGrid : matrix indexed by[STEPSEXCEPTLAST, INTERIORROWS, INTERIORCOLS]
  of int(0) union INTERIORROWS
\end{lstlisting}
The Boolean \code{matchingGrid} simply indicates whether the block at the specified location is involved in a match at this time step. The \code{mapsToGrid} behaves as a partial function, mapping the patterned block (if any) at the specified cell to its new destination in the next step, following the matching process and after the effects of gravity. This infrastructure is not needed in the final step: since no patterned blocks may remain, no matches can happen.


\subsubsection{Constraints: Initial and Goal States}
The initial and goal states are captured simply:
\begin{lstlisting}[escapechar=|, language=eprime]
$ Initial state:
forAll gCol : GRIDCOLS . forAll gRow : GRIDROWS . grid[0, gRow, gCol] = initGrid[gRow, gCol],

$ Goal state:
forAll gCol : GRIDCOLS . forAll gRow : GRIDROWS . grid[noSteps, gRow, gCol] <= EMPTY,
\end{lstlisting}
For the former we simply record the values in the parameter \code{initGrid} into the 0th time step of the \code{grid} state variables. For the latter, we insist that the final state contains only empty cells and wall blocks.

\subsubsection{Constraints: The Matching Grid}

The elements of \code{matchingGrid} at each time step are constrained as follows. Since the expression only covers the interior of the grid, and we assume the interior is surrounded by walls, there is no need to check for a boundary:
\begin{lstlisting}[escapechar=|, language=eprime]
forAll step : STEPSEXCEPTLAST .
  forAll gCol : INTERIORCOLS .
    forAll gRow : INTERIORROWS .
      ((grid[step, gRow, gCol] > EMPTY) /\
       ((grid[step, gRow-1, gCol] = grid[step, gRow, gCol]) \/ $North
        (grid[step, gRow+1, gCol] = grid[step, gRow, gCol]) \/ $South
        (grid[step, gRow, gCol+1] = grid[step, gRow, gCol]) \/ $East
        (grid[step, gRow, gCol-1] = grid[step, gRow, gCol])))  $West
      <-> (matchingGrid[step, gRow, gCol]),
\end{lstlisting}

Rather than introduce the \texttt{matchingGrid} variables, we could instead use the expression it is equivalent to, whenever it was needed.  This would naturally make the model more cumbersome to read, though not necessarily for Savile Row to process.

Our mode indicator variables, which controls whether a player action or the matching process is used to derive the state at the current time step, can then be constrained with respect to the contents of {\tt matchingGrid} at the previous time step:
\begin{lstlisting}[escapechar=|, language=eprime]
forAll step : STEPSFROM1 .
  matching[step] = (sum(flatten(matchingGrid[step-1,..,..])) > 0),
\end{lstlisting}

To avoid issues of symmetry, when in matching mode we constrain the player action variables to particular values, and, similarly, when in player action mode we force the values of the {\tt mapsToGrid} to be 0:
\begin{lstlisting}[escapechar=|, language=eprime]
forAll step : STEPSFROM1 .
  matching[step] ->
  ((moveRow[step] = 2) /\
   (moveCol[step] = 2) /\
   (moveDir[step] = -1)  /\
   (destRow[step] = 2) /\
   (destCol[step] = 2)),

forAll step : STEPSFROM1 .
  (!matching[step]) ->
  (forAll gCol : INTERIORCOLS .
    forAll gRow : INTERIORROWS .
      mapsToGrid[step-1,gRow,gCol] = 0),
\end{lstlisting}

\subsubsection{Constraints: The MapsTo Grid}

In matching mode, this grid is used to indicate the destination of each block in the previous state when deriving the next state. As noted above, \code{mapsToGrid} represents a partial function. Neither empty cells nor matched blocks, which will be removed from play, are mapped:
\begin{lstlisting}[escapechar=|, language=eprime]
$ Empty cells are not mapped
forAll step : STEPSFROM1 .
  forAll gCol : INTERIORCOLS .
    forAll gRow : INTERIORROWS .
      (grid[step-1,gRow,gCol] = EMPTY) ->
      (mapsToGrid[step-1,gRow,gCol] = 0),

$ Matched blocks are not mapped
forAll step : STEPSEXCEPTLAST .
  forAll gCol : INTERIORCOLS .
    forAll gRow : INTERIORROWS .
      (matchingGrid[step,gRow,gCol]) ->
      (mapsToGrid[step,gRow,gCol] = 0),
\end{lstlisting}
These two blocks of constraints need not be guarded by the \code{matching} mode indicator variables because they agree with the constraints to force \code{mapsToGrid} to 0 when not matching.

Wall blocks remain in place:
\begin{lstlisting}[escapechar=|, language=eprime]
forAll step : STEPSFROM1 .
  (matching[step]) ->
  (forAll gCol : INTERIORCOLS .
    forAll gRow : INTERIORROWS .
      grid[step-1,gRow,gCol] = WALL ->
      mapsToGrid[step-1,gRow,gCol] = gRow),
\end{lstlisting}

Unmatched pattern blocks must be mapped, and we constrain the mapping to preserve the ordering of these and the wall blocks in the resulting state:
\begin{lstlisting}[escapechar=|, language=eprime]
forAll step : STEPSFROM1 .
  (matching[step]) ->
  (forAll gCol : INTERIORCOLS .
    forAll gRow : INTERIORROWS .
      (!matchingGrid[step-1,gRow,gCol] /\
       grid[step-1,gRow,gCol] > EMPTY) ->
      (mapsToGrid[step-1,gRow,gCol] != 0)),

$ Bound how far a block can move
$ Triggers on both wall and blocks to ensure both types stay in right order.
forAll step : STEPSFROM1 .
  (matching[step]) ->
  (forAll gCol : INTERIORCOLS .
    forAll gRow : INTERIORROWS .
      (!matchingGrid[step-1,gRow,gCol] /\
       grid[step-1,gRow,gCol] != EMPTY) ->
      (forAll rowBeneath : int(gRow+1..gridHeight-1) .
        (!matchingGrid[step-1,rowBeneath,gCol] /\
         grid[step-1,rowBeneath,gCol] != EMPTY) ->
        (mapsToGrid[step-1,gRow,gCol] < mapsToGrid[step-1,rowBeneath,gCol]))),
\end{lstlisting}

\subsubsection{Constraints: Deriving the Next State when Matching}

In matching mode, we use the \code{mapsToGrid} partial function to derive the next state.
\begin{lstlisting}[escapechar=|, language=eprime]
$ Map the previous state to the new
forAll step : STEPSFROM1 .
  matching[step] ->
  forAll gCol : GRIDCOLS .
    forAll gRow : GRIDROWS .
      (mapsToGrid[step-1, gRow, gCol] > 0) ->
      (grid[step, mapsToGrid[step-1, gRow, gCol], gCol] = grid[step-1,gRow,gCol]),

$ Anything not part of the image of mapping function is empty
forAll step : STEPSFROM1 .
  (matching[step]) ->
  (forAll gCol : INTERIORCOLS .
    forAll gRow : INTERIORROWS .
     (forAll mRow : INTERIORROWS . mapsToGrid[step-1,mRow,gCol] != gRow)
     ->
     (grid[step,gRow,gCol] = EMPTY)),
\end{lstlisting}

We account for gravity simply by insisting that no block is floating:
\begin{lstlisting}[escapechar=|, language=eprime]
$ There are no floating blocks
forAll step : STEPSFROM0 .
  forAll gCol : INTERIORCOLS .
    forAll gRow : INTERIORROWS .
      (grid[step,gRow,gCol] > EMPTY) ->
      (grid[step,gRow+1,gCol] != EMPTY),
\end{lstlisting}

Finally, we deal with the border of wall blocks:
\begin{lstlisting}[escapechar=|, language=eprime]
forAll step : STEPSFROM1 .
  matching[step] ->
  forAll gCol : GRIDCOLS .
    forAll gRow : GRIDROWS .
      (((gCol = 1) \/ (gCol = gridWidth) \/ (gRow = 1) \/ (gRow = gridHeight))
       ->
       (grid[step,gRow,gCol] = grid[step-1,gRow,gCol])),
\end{lstlisting}

\subsubsection{Constraints: Moves by Player Action}

In player action mode, the move defined by the \code{move} and \code{dest} variables must be valid. We capture gravity in this account by constraining the \code{dest} variables to define location immediately above a non-empty cell and with a clear sequence of cells through which the selected block can fall:
\begin{lstlisting}[escapechar=|, language=eprime]
$ Select only valid blocks
forAll step : STEPSFROM1 .
  !matching[step]
  ->
  (grid[step-1, moveRow[step], moveCol[step]] > EMPTY),

$ destCol defined via moveDir
forAll step : STEPSFROM1 .
  !matching[step]
  ->
  (destCol[step] = moveCol[step] + moveDir[step]),

$ Can only move into clear space
forAll step : STEPSFROM1 .
  !matching[step]
  ->
  (grid[step-1, moveRow[step], destCol[step]] = EMPTY),
  
$ destRow must be at or below moveRow
forAll step : STEPSFROM1 .
  !matching[step]
  ->
  (destRow[step] >= moveRow[step]),
  
$ Something solid beneath destRow
forAll step : STEPSFROM1 .
  !matching[step]
  ->
  (grid[step-1, destRow[step]+1, destCol[step]] != EMPTY),

$ in destCol, everything from moveRow to destRow must be empty.
forAll step : STEPSFROM1 .
  !matching[step]
  ->
  (forAll gRow : GRIDROWS .
    ((gRow >= moveRow[step]) /\ (gRow <= destRow[step]))
    ->
    (grid[step-1, gRow, destCol[step]] = EMPTY)),
\end{lstlisting}

We capture the movement of the block selected by the player simply, using the \code{move} and \code{dest} variables:
\begin{lstlisting}[escapechar=|, language=eprime]
$ Contents of dest position are what was at source
forAll step : STEPSFROM1 .
  !matching[step]
  ->
  (grid[step-1, moveRow[step], moveCol[step]] =
   grid[step, destRow[step], destCol[step]]),
\end{lstlisting}

The action of moving a block from the source column might cause others above it to fall:
\begin{lstlisting}[escapechar=|, language=eprime]
forAll step : STEPSFROM1 .
  !matching[step]
  ->
  (forAll row : GRIDROWS .
    $ is affected by gravity?
    ((row <= moveRow[step]) /\
     (forAll rowBetween : int(row..gridHeight) .
       (rowBetween < moveRow[step])
       ->
       (grid[step-1, rowBetween, moveCol[step]] > EMPTY)
     ))
    ->
    ( $ Case 1: block falls into this row
     ((grid[step-1, row-1, moveCol[step]] > EMPTY) ->
      (grid[step, row, moveCol[step]] = grid[step-1, row-1, moveCol[step]])) /\
      $ Case 2: nothing to fall from above. This row becomes empty.
     ((grid[step-1, row-1, moveCol[step]] <= EMPTY) ->
      (grid[step, row, moveCol[step]] = EMPTY))
    )),
\end{lstlisting}

Finally, we must state frame axioms: constraints that maintain the parts of the state unaffected by the move.
\begin{lstlisting}[escapechar=|, language=eprime]
$ Frame axiom: Every column except source and destination remains the same.
forAll step : STEPSFROM1 .
  !matching[step]
  ->
  (forAll gCol : GRIDCOLS .
    ((gCol != moveCol[step]) /\ (gCol != destCol[step]))
    ->
    (forAll gRow : GRIDROWS . grid[step, gRow, gCol] = grid[step-1,gRow, gCol])),

$ Frame axiom: dest col - everything except destRow stays the same
forAll step : STEPSFROM1 .
  !matching[step]
  ->
  (forAll row : GRIDROWS .
    (row != destRow[step])
    ->
    (grid[step, row, destCol[step]] =
     grid[step-1, row, destCol[step]])),

$ Frame axiom: Source column - above moveRow unaffected by gravity stays same
forAll step : STEPSFROM1 .
  !matching[step]
  ->
  (forAll rowAbove : GRIDROWS .
    ((rowAbove < moveRow[step]) /\
     (exists rowBetween : int(rowAbove..gridHeight) .
       (rowBetween < moveRow[step]) /\
       (grid[step-1, rowBetween, moveCol[step]] <= EMPTY)
     ))
    ->
    (grid[step, rowAbove, moveCol[step]] =
     grid[step-1, rowAbove, moveCol[step]])),

$ Frame axiom: Source column - everything below moveRow[] stays the same
forAll step : STEPSFROM1 .
  !matching[step]
  ->
  (forAll rowBelow : GRIDROWS .
    (rowBelow > moveRow[step])
    ->
    (grid[step, rowBelow, moveCol[step]] = grid[step-1, rowBelow, moveCol[step]])),
\end{lstlisting}

\subsection{Variable Moves \texorpdfstring{\eprime}{Essence Prime} Model}

In this model we change focus from counting both moves and matching steps to counting only moves by the player. However, the Fixed Steps model described in \cref{section:fixedsteps} still underlies it.  We rename steps from the previous model to be `ministeps', with `steps' being reserved for player moves. An interesting point is that we can compute the maximum possible number of matching steps: each match removes at least two blocks, so the matching steps can be no more than half the number of patterned blocks in the original problem.

\begin{lstlisting}[escapechar=|, language=eprime]
letting maxMatches be (sum gRow : GRIDROWS . sum gCol : GRIDCOLS . (initGrid[gRow,gCol] > EMPTY)) /  2
letting noMiniSteps be noSteps + maxMatches
\end{lstlisting}

  Instead of insisting on an exact number of steps, we allow for `dummy' moves at the end of the sequence.   This means we have to compute the number of moves, and also add a statement to minimise the number of moves.  
  To avoid unnecessary symmetries we insist that all dummy moves are consecutive
  and that they leave all cells in the grid unchanged.  
\begin{lstlisting}[escapechar=|, language=eprime]
minimising numMoves
such that

numMoves = sum step : MINISTEPSFROM1 . moving[step],

forAll step : MINISTEPSFROM1 . 
    dummy[step] = !(matching[step] \/ moving[step]),

forAll step : int(1..noMiniSteps-1) . 
    dummy[step] -> dummy[step+1],

$ Dummy Mode Frame Axiom : all cells are unchanged
forAll step : MINISTEPSFROM1 .
  dummy[step]
  ->
  forAll gCol : GRIDCOLS . 
      forAll gRow : GRIDROWS . 
         grid[step-1,gRow,gCol] = grid[step,gRow,gCol],
\end{lstlisting}

The change to allow minimising number of moves has both advantages and disadvantages compared to the previous model.  The advantage is that we can use a single search to encompass a range of values of the number of moves, reducing the overhead of running Savile Row multiple times and conducting multiple searches.  The disadvantage is that if the given bound on moves is loose, then there is more overhead constructing the model and the search may be less efficient by having to find a number of solutions instead of just one.  

Note that while the number of steps is an upper bound instead of a precise number, we still have to supply a finite bound, as the number of variables in the model constructed by Savile Row depends on the bound.  In our experiments we adopted an exponential approach, doubling the bound each time until the selected maximum number.

We also use variables and constraints to summarise player moves, but omit these from this presentation as they should not affect search significantly.

\subsection{Potential Improvements to the Constraint Models}\label{sec:cp-improvements}

The models described in this section directly represent both the states and actions of the game, allowing us to straightforwardly place additional constraints on either states or actions (or both together). To further develop the models we can investigate breaking symmetries and dominances~\cite{Chu2015:dominance}, and adding implied constraints. The ability to extend a model in this way is a key advantage of constraint modelling compared to PDDL.


One approach to identify symmetries or dominances is to consider the sequence of states, ignoring the actions.  For example, if the same state occurs twice in a solution then a shorter solution exists; therefore duplicate states can safely be ruled out. 
Another approach would be to identify cases where the order of a pair of actions has no significance (i.e.\ executing the actions in either order leads to the same state). For example, suppose that a puzzle decomposes into two parts at step \(t\), and from \(t\) onwards the two parts are solved independently. In this case, any pair of adjacent moves that affect different parts of the puzzle would be interchangeable, and an order can be imposed to reduce search. 


Sometimes a lower bound can be found for the number of moves needed to clear the grid. As a simple example, suppose that there are two blocks remaining and the horizontal distance between them is \(k\). In this case, at least \(k-1\) moves are required to complete the puzzle. An implied constraint can be derived from any non-trivial lower bound, simply stating that there are sufficient moves remaining between the current time step and the horizon. 

Finally, it is possible to identify states that are not part of any solution. For example, if there is only one block remaining of any pattern, then the game cannot be solved. Analysis of the game rules may reveal other cases where the game becomes unsolvable, which can be ruled out by adding implied constraints (sometimes referred to as \textit{dead end} constraints~\cite{Espasa2022:plotting}).

\section{\texorpdfstring{\essence}{Essence} specification}

We further implemented a translation of the PDDL approach to the \essence language, making extensive use of \essence-specific features for conciseness. 

The \essence specification is a fairly faithful translation and uses the same directed graph point of view as the PDDL model: each reachable grid cell is represented by a vertex, labelled arcs \code{up}, \code{down}, \code{left}, or \code{right} link adjacent vertices, and walls are not present.

One difference between the \essence specification and the \eprime models described in \cref{section:constraints} is that the moves made by the player in the \essence specification are represented by a partial function assigning moves to some, but not necessarily all, of the integer time steps.
This avoids the need to deal with dummy moves.
A second difference is that, as with the PDDL model, falling actions are performed step by step rather than all at once.
This leads to long plans, with typically many states in which a single block falls from one cell to the one below it.
Matching actions are still performed all at once, as in the PDDL model.
Unlike the \eprime models, no attempt has been made to improve the efficiency of the \essence specification by means of collapsing multiple steps into one.

The main difference between the \essence specification and the PDDL model is the explicit scaffolding needed to support planning-style reasoning.
As in the \eprime models, in the \essence specification time is explicitly modelled as a sequence of discrete steps numbered by positive integers.
Actions are described as universally quantified formulas that link together the state at one time step with the state at the following step.
While the semantics of PDDL ensures that unchanged parts of the state space are implicitly copied across from one time step to the next,
the \essence specification needs to explicitly ensure that every part of the grid state at each time step is dealt with, and that cells that are not affected by any action at some time step stay the same into the next time step.

\begin{lstlisting}[escapechar=~, language=essence]
$ Essence specification for Puzznic without moving blocks, using POV of PDDL domain
letting patterns be new type enum {free, R, O, Y, G, B, V, P, C, L}
letting direction be new type enum {right, left, up, down}
given vertices new type enum
letting state be domain function (total) vertices --> patterns
$ arcs of adjacency relation on grid; walls are unreachable so are omitted
given A : relation of (vertices * vertices * direction) $ from, to, direction
   where forAll u : vertices . forAll d : direction . !A(u,u,d)
given initPat : function vertices --> patterns $ partial function
given horizon, maxMoves : int(1..)
letting time be domain int(1..horizon)
letting time1 be domain int(1..horizon-1)

find totalCost, done : time
find plan : function time --> state $ partial function
find falling, matching : function (total) time --> bool
find move : function time --> tuple(vertices,vertices) $ partial function
such that true
$ no dummy moves
,  forAll (u,v) in range(move) . u != v
,  forAll t in defined(plan) . ((t in defined(move)) <-> (!falling(t) /\ !matching(t) /\ (t < done)))

$ set up initial patterns; fill in free locations if needed
,  forAll u : vertices .
      ((u in defined(initPat)) -> (plan(1)(u) = initPat(u)))
   /\ ((!(u in defined(initPat))) -> (plan(1)(u) = free))
$ does something need to fall or match at the end of step t?
,  forAll t in defined(plan) . falling(t) =
      exists (u,v,d) in A . ((d=down) /\ (plan(t)(u) != free) /\ (plan(t)(v) = free))
,  forAll t in defined(plan) . matching(t) =
      exists (u,v,d) in A . ((plan(t)(u) = plan(t)(v)) /\ (plan(t)(u) != free))

$ all matches happen together
, forAll t : time1 . (matching(t) /\ !falling(t)) -> !matching(t+1)

$ a move is to a free adjacent L/R location, non-moving blocks remain
,  forAll t in defined(move) . and([ $ just for letting syntax
            (plan(t)(u) != free) /\ (plan(t)(v)  = free)
         /\ (A(u,v,left) \/ A(u,v,right))
         /\ (plan(t+1)(v) = plan(t)(u)) /\ (plan(t+1)(u) = free)
         | letting (u,v) be move(t) ])
,  forAll t in defined(move) . (
      forAll w : vertices . and([ ((w != u) /\ (w != v)) -> (plan(t+1)(w) = plan(t)(w))
      | letting (u,v) be move(t) ]) )

$ ensure a quiescent state once we are done
,  forAll t : time . ( (t >= done) -> (!matching(t) /\ !falling(t)) )

$ fall
,  forAll t : time1 . ( falling(t) $ postpone matches, if any
   -> forAll u : vertices . and([
         $ occupied and there is a free cell below it, so fall
         ((!freeU /\  fallUV) -> (plan(t+1)(u) = free)) $ from
         $ occupied, with no free cell below it, so keep
         /\ ((!freeU /\ !fallUV) -> (plan(t+1)(u) = plan(t)(u)))
         $ this is free, so the one above should fall here
         /\ (freeU -> (
               (forAll v : vertices . (A(v,u,down) -> (plan(t+1)(u) = plan(t)(v)))) $ to
            $ or if there isn't one above it, stay free
            /\ (!(exists v : vertices . A(v,u,down)) -> (plan(t+1)(u) = plan(t)(u)))))
      | letting freeU be (plan(t)(u) = free),
        letting fallUV be (exists v : vertices . (A(u,v,down) /\ (plan(t)(v)=free))) ]))

$ match (first fall then match)
,  forAll t : time1 . ( (!falling(t) /\ matching(t)) -> forAll u : vertices . (and([
         ( matchU -> (plan(t+1)(u) = free)) /\ (!matchU -> (plan(t+1)(u) = plan(t)(u)))
      | letting matchU be (exists (w,v,d) in A . (w=u)/\(plan(t)(u) = plan(t)(v))) ])))

$ goal: no patterned blocks left, within move budget
, forAll u : vertices . plan(done)(u) = free
,  totalCost = |defined(move)|
,  totalCost <= maxMoves
minimising done
\end{lstlisting}

This specification minimises \code{done}, the number of plan steps.
The move actions are straightforward (expressing the predicate \emph{this block moves either left or right into a free cell}), as is matching (expressing the predicate \emph{every occupied cell adjacent to another with the same pattern will match}).
The falling action is more intricate and uses locally scoped constants \code{freeU} (expressing the predicate \emph{this cell is free}) and \code{fallUV} (expressing \emph{there is a free cell below this one}).

\section{Experiments}

We used a set of benchmark instances consisting of levels from the PS1 version of the game, some levels from other versions of the game, and levels we constructed for debugging.

For the PDDL model, we used the Fast Downward planner~\cite{Helmert2006:fast}, release 22.12, with the blind search $A^{*}$ heuristic.
Since we were using the blind heuristic, we also disabled generation of negative axioms, as suggested by the author of Fast Downward\footnote{Personal communication, Malte Helmert, 2023.}.
Without this modification Fast Downward would typically exceed its compute budget during the generation of negative axioms, and therefore would fail to solve the problem, even for small instances.

To translate the \essence specification to \eprime, we used \conjure~\cite{Akgun2022:Conjure} version 2.4.1 with the default \code{-ac} compact heuristics.
For all \eprime constraint models, we used \savilerow~\cite{Nightingale2017:automatically} repository version e57ee1dc8 dated 2023-06-03, with the \code{-sat-polarity} option.
The backend SAT solver used was Kissat version 3.0.0~\cite{cadical} with default options.
Translation time was generally negligible compared to the time taken by the SAT solver to obtain a solution.

\zap{
\joan{
Experimental detail:
The axioms version of the model seems to work well when disabling the computation of negative axioms. That is, builds/release/bin/translate/axiom\_rules.py:67.

According to Malte Helmert, some heuristics that use that might become unsafe, but using the blind heuristic should be fine: \url{https://discord.com/channels/800677597036937256/974644163888971786/1095998141045686303}

Mateu's example shows that the PDDL model does not really fit into the semantics we expect. It finds the shorter way to explode everything.
}
}

\begin{table}[p]
\begin{tabular}{@{}l|cr|cr|cr||cr}
Instance & \multicolumn{2}{c|}{Fixed Steps} & \multicolumn{2}{c}{Minimising Moves} & \multicolumn{2}{|c||}{\essence} & \multicolumn{2}{c}{PDDL} \\
& opt? & cpu(s) & opt? & cpu(s) & opt? & cpu(s) & opt? & cpu(s) \\
\hline
5x7-ps1-a13 & Y & 35.72 & Y & {\bf 15.65}
& Y & 37.18
& Y & 0.23
\\
6x6-mateu & Y & 9.48 & Y & {\bf 8.84}
& Y & 47.43
& Y & 0.25
\\
6x6-ps1-c28 & Y & 58.94 & Y & {\bf 25.44}
& Y & 42.27
& Y & 0.23
\\
6x7-ipg & Y & 9.64 & Y & {\bf 8.68}
& Y & 43.84
& Y & 0.29
\\
6x7-mateu & Y & {\bf 5.80} & Y & 11.85
& Y & 42.05
& Y & 0.29
\\
6x9-mateu & Y & {\bf 8.00} & Y & 19.38
& Y & 64.26
& Y & 0.43
\\
6x11-bcl-015-5 & N & $>3600$ & N & $>3600$
& U & {\bf 40.98} 
& U & 0.55
\\
7x7-mateu & Y & 37.42 & Y & {\bf 12.42}
& Y & 74.11
& Y & 0.38
\\
8x11-ps1-e11 & Y & 1265.35 & Y & {\bf 1248.00}
& N & $>7200$ 
& Y & 8.42
\\
8x12-ps1-e22 & Y & {\bf 990.27} & N &  $> 3600$
& Y & 4123.65
& Y & 1.45
\\
8x12-ps1-e47 & Y & {\bf 3136.92} & N &  $> 3600$
& N & $>7200$ 
& Y & 106.19
\\
8x7-mateu & Y & 53.58 & Y & {\bf 16.47}
& Y & 84.92
& Y & 0.48
\\
8x7-ps1-d21 & N & $>3600$ & N & $>3600$
& U & {\bf 759.28}
& U & 0.45
\\
8x8-ps1-b15 & Y & 196.50 & Y & 232.32
& Y & {\bf 131.53}
& Y & 0.81
\\
8x8-ps1-d18 & Y & {\bf 317.63} & Y & 501.16
& N & $>7200$
& Y & 0.82
\\
8x8-ps1-e35 & N & $>3600$ & N & $>3600$
& N & $>7200$
& U & 76.74
\\
9x5-ps1-b22 & Y & 124.14 & Y & 55.67
& Y & {\bf 48.51}
& Y & 0.29
\\
9x7-cubic & N & $>3600$ & N & $>3600$
& N & $>7200$
& Y & 8.40
\\
9x7-ps1-b12 & Y & 452.18 & Y & 1297.09
& Y & {\bf 256.28}
& Y & 0.68
\\
9x12-ps1-c17 & Y & {\bf 457.48} & Y & 2681.86
& N & $>7200$
& Y & 10.86
\\
10x7-bcl-014-2 & N & $>3600$ & N & $>3600$
& Y & 99.27
& Y & 0.63
\\
10x12-ps1-d23 & Y & {\bf 437.77} & Y & 970.35
& M & - 
& Y & 3.04
\\
10x12-ps1-d27 & Y & {\bf 3282.71} & N &  $> 3600$
& Y & 4450.85
& Y & 2.32
\\
10x12-ps1-e21 & Y & {\bf 1462.20} & N &  $> 3600$
& M & - 
& Y & 385.42
\\
10x12-ps1-e28 & N & $>3600$ & N & $>3600$
& N & $>7200$
& Y & 648.44
\\
10x5-ps1-b11 & Y & 213.46 & Y & 84.03
& Y & {\bf 70.15}
& Y & 0.44
\\
10x7-ps1-c16 & Y & {\bf 42.82} & Y & 78.65
& N & $>7200$
& Y & 14.03
\\
10x7-ps1-d34 & Y & {\bf 620.06} & Y & 743.18
& N & $>7200$
& Y & 9.35
\\
10x8-bip-001-42 & N & $>3600$ & N & $>3600$
& N & $>7200$
& Y & 1504.66
\\
10x8-bv1-001-12 & Y & 378.85 & Y & 1073.07
& Y & {\bf 292.15}
& Y & 0.98
\\
10x12-ps1-e45 & N & $>3600$ & N & $>3600$
& N & $>7200$
& Y & 1979.51
\\
10x11-ps1-e48 & N & $>3600$ & N & $>3600$
& N & $>7200$
& Y & 3.62
\\
%
\zap{ 
5x100-test & & & &
& M & - 
& Y & 24.11
\\
10x12-ps1-e24 & N & $>3600$ & N & $>3600$
& N & $>7200$
& M & - 
\\
10x12-ps1-e31 & & & &
& N & $>7200$
& N & $>7200$
\\
10x12-ps1-d22
& N & $>7200$
& N & $>7200$
\\
10x12-ps1-d35 & & & &
& N & $>7200$
& N & $>7200$
\\
10x12-ps1-e17 & & & &
& N & $>7200$
& N & $>7200$
}
\\ 
\end{tabular}
\caption{In column opt?, Y indicates that the method was able to find a solution and prove it optimal (on its measure) within its CPU time limit (2 hours for PDDL and \essence, 1 hour for \eprime).
The CPU time indicated includes all time reported by both Savile Row in its modelling phase and the time used by the solver, over all individual instances solved to find the proof.
U indicates that the instance was shown to be unsolvable (within a 100 move horizon for the non-PDDL approaches), N that the method was unable to prove either optimality of a solution or unsolvability, and M that the method exceeded memory bounds (in this case the time taken is not meaningful).
For five instances (\code{10x12-ps1-\{d22,d34,e17,e24,e31\}}) every solver timed out or exceeded memory limits, and these are not shown in the table.
PDDL dominates all instances, but we have highlighted in bold the fastest constraint approach in each case.
}
\label{table:results}
\end{table}

We summarise our experiments in \cref{table:results}.
The PDDL approach outperforms our current constraint programming approaches.
However, as previously discussed, we had to disable the generation of negative axioms in Fast Downward to avoid the solver timing out.

None of our approaches is able to solve more complex instances, including several levels from the PS1 version of Puzznic which are not especially challenging for human players but which cause every approach to time out or exceed memory bounds.
The \code{10x12-ps1-e21} instance was notable as being quite challenging for the PDDL approach while still being solved. For this instance the Fixed Steps model achieved reasonable performance, the Variable Steps approach timed out, and the \essence specification exceeded memory bounds.

The \eprime and \essence approaches have different performance characteristics.
Some instances (such as \code{10x8-bv1-001-12}) are solved faster with the \essence approach, while for many of our instances the \eprime approaches are faster.

The Minimising Moves version of the Variable Steps \eprime model sometimes outperforms the Fixed Steps model, and vice versa. However, the Fixed Steps model usually performs better, and the Variable Steps model never solves an instance within the timeout when the Fixed Steps model fails to do so.

The \essence specification fails for large instances, as \savilerow runs out of memory when the graph of cells contains more than about 220 vertices.
Inserting many empty rows at the top of the grid does not change possible solutions but reliably triggers this failure.

\section{Summary and Further Work}

The video game Puzznic presents an interesting modelling challenge.
We have modelled Puzznic (without moving blocks) using both constraint programming and planning tools. For the constraint approaches, we used both \essence and \eprime, with the modelling tools \conjure~\cite{Akgun2022:Conjure} and \savilerow~\cite{Nightingale2017:automatically}, and in each case the models were translated to SAT and solved with Kissat~\cite{cadical}.
In planning, we used PDDL and the Fast Downward planner~\cite{Helmert2006:fast}. In all, we compare four approaches to modelling and solving Puzznic on a small set of benchmark game levels.

Comparisons between our approaches are complicated by the fact that the found plans are of minimal length but the definition of length differs (e.g.\ the number of game events vs player actions). The planning approach is at present superior to our constraint programming approaches. For these approaches, we found that the Fixed Steps model is usually solved faster than the Minimising Moves model, and mostly also outperforms the \essence approach.

We have already observed large improvements in solution time by improving our constraint models, and continue to improve our models.
In future work, we will explore symmetries and dominances, bounds on the number of moves required, and dead end constraints as described in \Cref{sec:cp-improvements}. 
We also intend to preprocess instances to remove grid cells which are not required to obtain a successful plan, and to reduce the complexity of the \eprime models and the \essence specification to allow \savilerow to more easily digest such instances.
We will also investigate the differences in performance between the \eprime models and the \essence specification.
Finally, we will consider instance generation and prepare a larger set of benchmark instances to contrast strengths and weaknesses of different approaches to solving this problem.

\newpage


\bibliography{main,general}

\begin{thebibliography}{10}

\bibitem{Akgun2022:Conjure}
{\"O}zg{\"u}r Akg{\"u}n, Alan~M. Frisch, Ian~P. Gent, Christopher Jefferson,
  Ian Miguel, and Peter Nightingale.
\newblock Conjure: Automatic generation of constraint models from problem
  specifications.
\newblock {\em Artificial Intelligence}, 310:103751, 2022.
\newblock \href {https://doi.org/10.1016/j.artint.2022.103751}
  {\path{doi:10.1016/j.artint.2022.103751}}.

\bibitem{bartak2010constraint}
Roman Bart{\'a}k, Miguel~A Salido, and Francesca Rossi.
\newblock Constraint satisfaction techniques in planning and scheduling.
\newblock {\em Journal of Intelligent Manufacturing}, 21(1):5--15, 2010.

\bibitem{related1}
Roman Bart{\'{a}}k and Daniel Toropila.
\newblock Reformulating constraint models for classical planning.
\newblock In {\em Proceedings of the Twenty-First International Florida
  Artificial Intelligence Research Society Conference, May 15-17, 2008}, pages
  525--530. {AAAI} Press, 2008.

\bibitem{cadical}
Armin Biere, Katalin Fazekas, Mathias Fleury, and Maximillian Heisinger.
\newblock {CaDiCaL}, {Kissat}, {Paracooba}, {Plingeling} and {Treengeling}
  entering the {SAT Competition 2020}.
\newblock In {\em Proceedings of {SAT} {Competition} 2020 -- Solver and
  Benchmark Descriptions}, pages 50--53, 2020.
\newblock URL: \url{http://hdl.handle.net/10138/318754}.

\bibitem{chavrimootoo2022:defying}
Michael~C. Chavrimootoo.
\newblock Defying gravity: The complexity of the {Hanano} puzzle.
\newblock arXiv, 2022.
\newblock \href {https://doi.org/10.48550/arXiv.2205.03400}
  {\path{doi:10.48550/arXiv.2205.03400}}.

\bibitem{Chu2015:dominance}
Geoffrey Chu and Peter~J. Stuckey.
\newblock Dominance breaking constraints.
\newblock {\em Constraints}, 20(2):155--182, 2015.
\newblock \href {https://doi.org/10.1007/s10601-014-9173-7}
  {\path{doi:10.1007/s10601-014-9173-7}}.

\bibitem{Espasa2022:plotting}
Joan Espasa, Ian Miguel, and Mateu Villaret.
\newblock Plotting: a planning problem with complex transitions.
\newblock In {\em 28th International Conference on Principles and Practice of
  Constraint Programming (CP 2022)}, pages 22:1--22:17, 2022.
\newblock \href {https://doi.org/10.4230/LIPIcs.CP.2022.22}
  {\path{doi:10.4230/LIPIcs.CP.2022.22}}.

\bibitem{cubicnp}
Erich Friedman.
\newblock The game of cubic is {NP}-complete.
\newblock 34th Annual Florida MAA Section Meeting, 2001.

\bibitem{Gent2007:search}
Ian~P Gent, Chris Jefferson, Tom Kelsey, In{\^e}s Lynce, Ian Miguel, Peter
  Nightingale, Barbara~M Smith, and S~Armagan Tarim.
\newblock Search in the patience game ‘black hole’.
\newblock {\em AI Communications}, 20(3):211--226, 2007.
\newblock \url{https://content.iospress.com/articles/ai-communications/aic405}.

\bibitem{ghallab2004automated}
Malik Ghallab, Dana Nau, and Paolo Traverso.
\newblock {\em Automated Planning: theory and practice}.
\newblock Elsevier, 2004.

\bibitem{pddl}
Patrik Haslum, Nir Lipovetzky, Daniele Magazzeni, and Christian Muise.
\newblock {\em An Introduction to the Planning Domain Definition Language}.
\newblock Synthesis Lectures on Artificial Intelligence and Machine Learning.
  Springer, 2019.
\newblock \href {https://doi.org/10.2200/S00900ED2V01Y201902AIM042}
  {\path{doi:10.2200/S00900ED2V01Y201902AIM042}}.

\bibitem{Helmert2006:fast}
Malte Helmert.
\newblock The {Fast} {Downward} planning system.
\newblock {\em Journal of Artificial Intelligence Research}, 26:191--246, 2006.
\newblock \href {https://doi.org/10.1613/jair.1705}
  {\path{doi:10.1613/jair.1705}}.

\bibitem{Jefferson2006:modelling}
Christopher Jefferson, Angela Miguel, Ian Miguel, and Armagan Tarim.
\newblock Modelling and solving {English} {Peg} {Solitaire}.
\newblock {\em Comput. Oper. Res.}, 33(10):2935--2959, 2006.
\newblock \href {https://doi.org/10.1016/j.cor.2005.01.018}
  {\path{doi:10.1016/j.cor.2005.01.018}}.

\bibitem{Liu2019:hanano}
Ziwen Liu and Chao Yang.
\newblock {Hanano} puzzle is {NP}-hard.
\newblock {\em Information Processing Letters}, 145:6--10, 2019.
\newblock \href {https://doi.org/10.1016/j.ipl.2019.01.003}
  {\path{doi:10.1016/j.ipl.2019.01.003}}.

\bibitem{savile-row-manual-caps}
Peter Nightingale.
\newblock Savile {Row} manual, 2021.
\newblock \href {https://doi.org/10.48550/arXiv.2201.03472}
  {\path{doi:10.48550/arXiv.2201.03472}}.

\bibitem{Nightingale2017:automatically}
Peter Nightingale, {\"O}zg{\"u}r Akg{\"u}n, Ian~P. Gent, Christopher Jefferson,
  Ian Miguel, and Patrick Spracklen.
\newblock Automatically improving constraint models in {Savile Row}.
\newblock {\em Artificial Intelligence}, 251:35--61, 2017.
\newblock \href {https://doi.org/10.1016/j.artint.2017.07.001}
  {\path{doi:10.1016/j.artint.2017.07.001}}.

\end{thebibliography}

\end{document}